\pdfoutput=1

\documentclass[11pt]{article}

\usepackage[final]{acl}

\usepackage{times}
\usepackage{latexsym}

\usepackage[T1]{fontenc}

\usepackage[utf8]{inputenc}

\usepackage{microtype}

\usepackage{inconsolata}

\usepackage{graphicx}

\usepackage{amsmath}
\usepackage{amssymb}
\usepackage{booktabs}
\usepackage{multirow}

\title{Beyond Input Activations: Identifying Influential Latents by \\Gradient Sparse Autoencoders}

\author{Dong Shu\textsuperscript{1,*}, Xuansheng Wu\textsuperscript{2,}\thanks{These authors contributed equally.}, \textbf{Haiyan Zhao\textsuperscript{3}},  \textbf{Mengnan Du\textsuperscript{3}}, \textbf{Ninghao Liu\textsuperscript{2}}\\
\textsuperscript{1}Northwestern University \,
\textsuperscript{2}University of Georgia \,\\
\textsuperscript{3}New Jersey Institute of Technology\\
\small\texttt{dongshu2024@u.northwestern.edu}, \small\texttt{\{xw54582,ninghao.liu\}@uga.edu}, \small\texttt{\{hz54,mengnan.du\}@njit.edu}
}

\begin{document}
\maketitle

\begin{abstract}
Sparse Autoencoders (SAEs) have recently emerged as powerful tools for interpreting and steering the internal representations of large language models (LLMs). However, conventional approaches to analyzing SAEs typically rely solely on input-side activations, without considering the causal influence between each latent feature and the model's output. This work is built on two key hypotheses: (1) activated latents do not contribute equally to the construction of the model's output, and (2) only latents with high causal influence are effective for model steering. To validate these hypotheses, we propose \textbf{Gradient Sparse Autoencoder} (\textbf{GradSAE}), a simple yet effective method that identifies the most influential latents by incorporating output-side gradient information. Our code is available at \url{https://github.com/Tizzzzy/sae_gradient}.


\end{abstract}

\begin{figure*}
    \centering
    \includegraphics[width=0.90\linewidth]{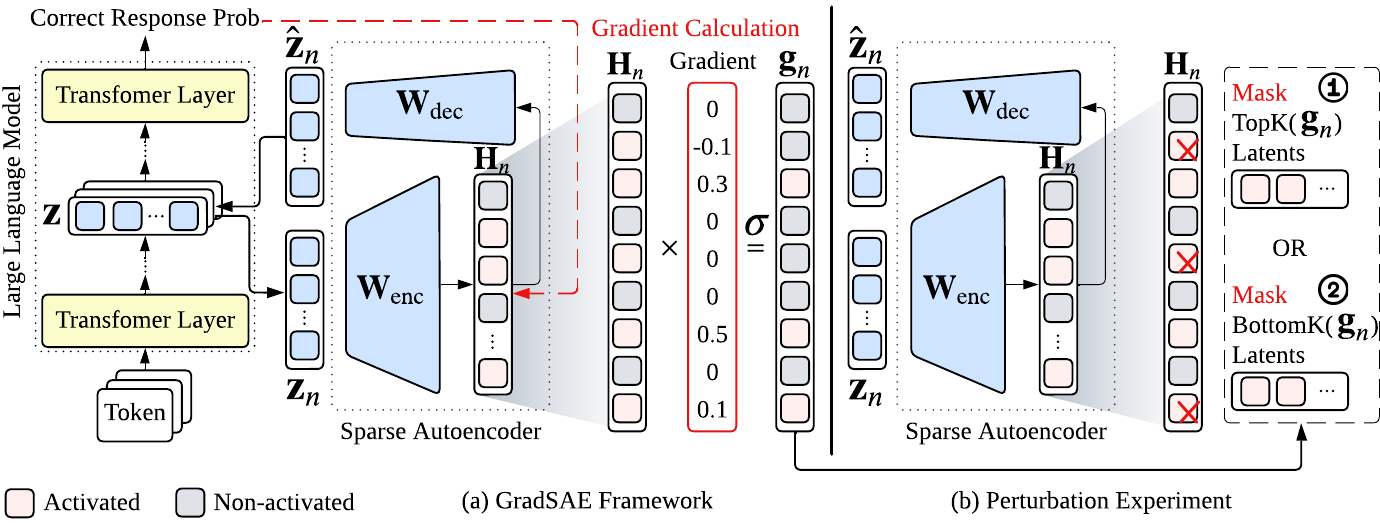}
    \caption{Methodology overview in the perturbation experiment. Subfigure (a) illustrates the GradSAE framework, where the symbol $\sigma$ denotes the ReLU function. Subfigure (b) shows the experiment process, which contains two settings and both share the same architecture with the GradSAE framework but differ in masking strategies.}
    \label{fig:overflow}
    \vspace{-10pt}
\end{figure*}
\vspace{-0.3cm}
\section{Introduction}
\vspace{-0.1cm}
Sparse Autoencoders (SAEs) have recently emerged as promising tools for interpreting the inner workings of large language models (LLMs) \cite{cunningham2023sparse, bricken2023monosemanticity, gao2025scaling, rajamanoharan2024jumping}. A core challenge in understanding LLMs is the polysemanticity of neurons, where each neuron encodes multiple features \cite{arora2018linear, scherlis2022polysemanticity}. This is largely due to superposition \cite{elhage2022superposition}, a phenomenon where the number of features an LLM needs to represent vastly exceeds the number of available neurons. SAEs address this by learning an overcomplete latent space, allowing each \textit{latent} to represent a single, disentangled feature. For any given LLM representation, only a small number of these latents are activated, and the combination of these sparse active latents can accurately reconstruct the original LLM representation. This sparsity makes it easier to interpret the concepts an LLM is processing.

Although interpretability was the original motivation for developing SAEs, they have proven useful for other applications as well, particularly in \textit{steering model behaviors} \cite{chalnev2024improving, he2025saif, zhao2024steering, galichin2025have}. Traditionally, researchers associate each latent with a human-interpretable concept by analyzing which input texts tend to activate it. By modifying selected latents in the SAE space that have desired concepts, ideally we can influence the LLM outputs toward our expectation in a controllable way \cite{templeton2024scaling, o2024steering}. However, these approaches assume that \textbf{the latent's activation based on input reflects a causal influence on the model output, which has never been proven.} Recent evidence suggests that this assumption may not always hold, and steering can sometimes produce unintended effects on the output \cite{durmus2024steering,wu2025interpreting}.


In this paper, we argue that identifying latents solely from input activations is insufficient for reliable model steering. Instead, the relationship between SAE latents and LLM output should also be considered when determining which latents are most relevant for intervention. To address this, we propose Gradient SAE (GradSAE), a simple yet effective method that can be applied to any instruction-tuned LLM's SAE. 
Our key insight is that \textbf{not all latents activated by the input contribute equally to generating the model's output.} Rather, only those latent variables whose activations, when set to zero, lead to a significant change in the model's outputs are likely to exert substantial influence. In our paper, we prove that this ablation process can be approximated with a more efficient gradient-based approach. To validate our hypothesis, we design two experiments. First, we demonstrate that activated latents have different impacts when used to generate model outputs. Second, we show that the influential latents, identified by GradSAE, are more effective for output steering.

\section{Methodology}
\subsection{Problem Statement}

Let \( \mathcal{V} \) be the vocabulary, \( \mathcal{X} \in \mathcal{V}^N \) an input sequence, and \( \mathcal{Y} \in \mathcal{V}^M \) the corresponding LLM-generated output. The hidden representation at layer \( l \) is \( \mathbf{Z}^{(l)} \in \mathbb{R}^{N \times D} \), where \( D \) is the hidden dimension. 
We omit the superscript $^{(l)}$ for simplicity in the rest of the paper. 
A \textit{pre-trained} SAE is inserted at layer $l$ with parameters $\mathbf{W}_\text{enc}\in\mathbb{R}^{D\times C}$ and $\mathbf{W}_\text{dec}\in\mathbb{R}^{C\times D}$, where $C$ is the latent dimension of SAE and $C\gg D$. 
Given the representation $\mathbf{Z}$ of the entry $\mathcal{X}$, SAE first decomposes $\mathbf{Z}$ as a sparse latent activation $\mathbf{H}\in\mathbb{R}^{N\times C}$, and then restores $\hat{\mathbf{Z}}\in\mathbb{R}^{N\times D}$ with $\mathbf{H}$ as:
\begin{align}
\hat{\mathbf{Z}} &= \mathbf{H}\mathbf{W}_{\text{dec}}=\sigma(\mathbf{Z}\mathbf{W}_{\text{enc}})\mathbf{W}_{\text{dec}}.
\end{align}
Here, $\sigma$ is a non-linear activation function, and $\hat{\mathbf{Z}}$ is subsequently passed to the rest of layers. 
We aim to identify which learned latent $c$ in \( \mathbf{H}\) is most causally influential in generating \( \mathcal{Y} \). 

\subsection{Proposed GradSAE Framework}
\label{sec:gradsae_framework}
We illustrate our proposed GradSAE framework for estimating the influences of sparse latent activations $c=1,...,C$ in Figure~\ref{fig:overflow}a. 
Following previous work~\cite{feng2018pathologies,wu2024language}, we define the influence of a certain latent $c$ at the $n$-th input token on the output $\mathcal{Y}$ as the change in prediction with and without sparse latent activation $\mathbf{H}_{n,c}\in\mathbb{R}$:
\begin{equation}
    \mathbf{g}_{n,c}=p(\mathcal{Y}|\mathbf{H}) - p(\mathcal{Y}|\mathbf{H}_{n,/c}),
    \label{eq:influence}
\end{equation}
where $\mathbf{H}_{n,/c}$ indicates setting the $c$-th value of the $n$-th row at $\mathbf{Z}$ as 0, and probability $p(\mathcal{Y}|\cdot)$ is predicted logits of LLM on $\mathcal{Y}$ with original $\mathbf{H}$ or masked $\mathbf{H}_{n,/c}$ sparse latent activations of SAE.

For efficiency, we approximate $\mathbf{g}_{n,c}$ in Equation~\eqref{eq:influence} with the gradients of output logit respecting to the latent activations $\mathbf{Z}_{n,c}$ at the $n$-th input token (see proof in Appendix~\ref{appendix:proof}). Let \( \mathbf{H} = \{ \mathbf{H}_1, ..., \mathbf{H}_N \} \) represent token-wise latent activations of the input sequence. Thus, the influence of the $c$-th latent activation on the $n$-th token is:
\begin{equation}
\mathbf{g}_{n,c} \approx \frac{\partial p(\mathcal{Y} | h(\mathbf{Z)})}{\partial \mathbf{H}_{n,c}} \odot \mathbf{H}_{n,c},
\label{eq:impirical}
\end{equation}
where $\odot$ indicates element-wise multiplication. In practice, we only focus on the latents that show positive influences to outputs. 
Equation~\eqref{eq:impirical} reveals that the magnitude of the raw sparse latent activation (i.e., $\mathbf{H}_{n,c}$) along cannot effectively estimate its influence to LLM outputs, while many existing works~\cite{templeton2024scaling,o2024steering} regardless this fact and simply use the latent activations to interpret SAEs and/or steer LLMs. 
We finally define the overall influence on the $c$-th latent by average individual influence scores across the input sequence, i.e., $\mathbf{g}_c=\frac{1}{N}\sum_{n=1}^N \mathbf{g}_{n,c}$.

\section{Experiments}
We empirically investigate the following research questions (RQs).
\textbf{RQ1}: Do all activated latents contribute equally to construct the model's output?
\textbf{RQ2}: How effectively does GradSAE identify the latents that significantly influence the model's output?
\textbf{RQ3}: Can the latents selected by GradSAE lead to better output steering?

\subsection{General Settings}
\subsubsection{Dataset and Metrics}
In this paper, we use the SQuAD dataset~\cite{rajpurkar2016squad}, where each example consists of a context passage, a question, and an answer. We adopt the standard SQuAD evaluation metrics: Exact Match (EM) and token-level F1. Detailed dataset statistics and description are provided in Appendix \ref{appendix:dataset}, and metric descriptions are in Appendix \ref{appendix:metrics}.

\begin{table*}[t]
\centering
\caption{Results of both perturbation and local steering. The upper section shows perturbation results, where for the TopK rows, lower scores indicate greater influence on the model output, and for the BottomK rows, higher scores indicate less influence. The best performance in each column is in \textbf{bold}. The lower section shows local steering results, where for the TopK rows, higher scores indicate stronger steering effects on the model output, and for the BottomK rows, lower scores indicate weaker steering effects. The highest score in each column is in \underline{underlined}.}
\scalebox{0.7}{
\label{tab:first_experiment}
\begin{tabular}{c|ccccccccccccc}
\toprule
\toprule
\multirow{2}{*}{Tasks} &  & & \multirow{2}{*}{w/o Task} & \multicolumn{2}{c}{K=1} & \multicolumn{2}{c}{K=10} & \multicolumn{2}{c}{K=20} & \multicolumn{2}{c}{K=30} & \multicolumn{2}{c}{K=50\%} \\ \cline{5-14}
 &  & &  & EM & F1 & EM & F1 & EM & F1 & EM & F1 & EM & F1 \\ \hline
\multirow{4}{*}{Perturbation} 
 & \multirow{2}{*}{Baseline} & TopK \textcolor{blue}{$\downarrow$} & \multirow{2}{*}{100.0} & 94.79  & 97.04  & 91.81 & 94.46  & 85.47 & 88.49 & 75.42 & 79.62 & 53.42 & 58.46 \\
 &  & BottomK \textcolor{blue}{$\uparrow$} &  & 100.0 & 100.0 & 98.51 & 99.93  & 98.51 & 99.93 & 98.51 & 99.93 & 98.41 & 99.84 \\ \cline{2-14}
 & \multirow{2}{*}{GradSAE} & TopK \textcolor{blue}{$\downarrow$} & \multirow{2}{*}{100.0} & \textbf{80.45} & \textbf{82.67}  & \textbf{51.21} & \textbf{60.46}  & \textbf{39.48} & \textbf{50.51} & \textbf{37.80} & \textbf{49.37} & \textbf{30.58} & \textbf{43.33} \\
 &  & BottomK \textcolor{blue}{$\uparrow$} & & 100.0 & 100.0  & 98.14  & 99.68  & 98.14 & 99.68 & 98.14 & 99.68 & 98.21 & 99.66 \\ \hline \hline
\multirow{4}{*}{Local Steering} 
 & \multirow{2}{*}{Baseline} & TopK \textcolor{red}{$\uparrow$} & \multirow{2}{*}{0.00} & 4.18 & 5.77  & 4.59 & 6.69  & 2.19 & 3.32 & 1.00 & 1.62 & 0.20 & 0.55 \\
 &  & BottomK \textcolor{red}{$\downarrow$} & & 0.00  & 0.00  & 0.00  & 0.00  & 0.00  & 0.00  & 0.00  & 0.00  & 0.00  & 0.00  \\ \cline{2-14}
 & \multirow{2}{*}{GradSAE} & TopK \textcolor{red}{$\uparrow$} & \multirow{2}{*}{0.00} & \underline{4.58} & \underline{6.43} & \underline{7.99} & \underline{10.21}  & \underline{4.78} & \underline{6.63}  & \underline{3.59} & \underline{4.88} & \underline{3.39} & \underline{4.58} \\
 &  & BottomK \textcolor{red}{$\downarrow$} &  & 0.00  & 0.00  & 0.00  & 0.00  & 0.00  & 0.00  & 0.00  & 0.00  & 0.00  & 0.00 \\ 
\bottomrule
\bottomrule
\end{tabular}}
\vspace{-10pt}
\end{table*}

\subsubsection{Perturbation Experiment (RQ1 \& RQ2)}
\label{sec:first_experiment_design}

\paragraph{Experimental Designs.} As shown in Figure~\ref{fig:overflow}b, our GradSAE experiments involve two settings. Before conducting these, we define two sets of latents:
\begin{align}
\vspace{-0.4cm}
\mathcal{Z}_{\text{high}}&=\text{argmax}_{\mathcal{Z}^\prime \subset \mathcal{Z}_{\text{NZ}},|\mathcal{Z}^\prime|=K}\sum_{c\in \mathcal{Z}^\prime} \mathbf{g}_c, \\
\mathcal{Z}_{\text{low}}&=\text{argmin}_{\mathcal{Z}^\prime \subset \mathcal{Z}_{\text{NZ}},|\mathcal{Z}^\prime|=K}\sum_{c\in \mathcal{Z}^\prime} \mathbf{g}_c, 
\vspace{-0.4cm}
\end{align}
where \( \mathcal{Z}_{\text{NZ}} = \{ c \mid \mathbf{g}_c > 0, c=1,..,C \} \) are the indices of positive latent influences. In our experiments, we define \( K \in \{1, 10, 20, 30, 50\%\} \), where ``50\%'' denotes half the number of non-zero latents count in the SAE activation of the last token. 



Let \( \mathbf{H} \in \mathbb{R}^{N \times C} \) denote the sequence of token-wise activations. In the first setting, for each token's activation vector, we zero out all latents in \( \mathcal{Z}_{\text{high}} \). In the second setting, we instead zero out the latents in \( \mathcal{Z}_{\text{low}} \). The modified activation is then passed through \( \mathbf{W}_{\text{dec}} \) to reconstruct \( \hat{\mathbf{Z}} \) and resume the forward pass. If masking \( \mathcal{Z}_{\text{high}} \) leads to degraded performance while masking \( \mathcal{Z}_{\text{low}} \) has no impact, this supports our hypothesis that not all activated latents contribute equally to the model's output. Note that we focus only on samples where the LLM can correctly answer with greedy decoding, resulting in 100\% accuracy without any perturbation. If masking \( \mathcal{Z}_{\text{low}} \) truly has no impact, the performance after perturbation should remain close to 100\%.

\paragraph{Baseline.} We repeat the GradSAE framework in Section \ref{sec:gradsae_framework} but without using gradient information. Specifically, we compute the mean over the original token-wise SAE activations, skipping the gradient calculation:
$\mathbf{g}_c=\frac{1}{N}\sum_{n=1}^N \mathbf{H}_{n,c}$
We then extract \( \mathcal{Z}_{\text{high}} \) and \( \mathcal{Z}_{\text{low}} \) from this baseline vector and compare performance when masking these sets. 

\begin{figure}
    \centering
    \includegraphics[width=0.75\linewidth]{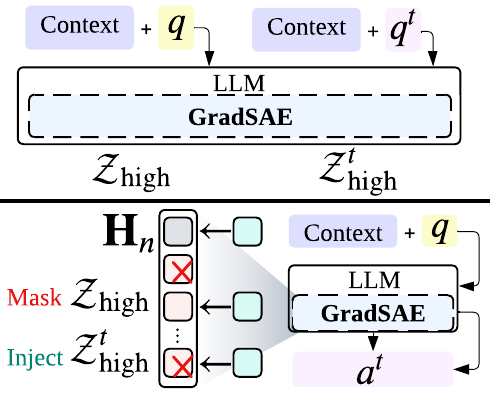}
    \caption{Local steering methodology overview.}
    \vspace{-0.2cm}
    \label{fig:steer_experiment_main}
    \vspace{-15pt}
\end{figure}

\subsubsection{Local Steering Experiment (RQ3)}
This experiment aims to address \textbf{RQ3} following a similar design to the perturbation experiment (Section~\ref{sec:first_experiment_design}). The key difference in the steering experiment is that, when extracting the TopK and BottomK latents, we also extract the corresponding value. This allows us to adjust the activations using these values in the subsequent settings.

\paragraph{Experimental Designs.} We consider a particular steering task, called local steering, leveraging a unique property of the SQuAD dataset: each context passage is paired with multiple questions. This allows us to investigate whether the model's output for one question can be steered by latent activations derived from a different question that shares the same context. As shown in Figure \ref{fig:steer_experiment_main} upper part, for each data point \( d = (\text{context}, q, a) \in D \), we define a set of examples with the same context but different questions as \( D^{\text{same}} = \{ d^t = (\text{context}, q^t, a^t) \mid q^t \ne q \} \). We randomly select one such example $ d^t \in D^{\text{same}} $, and denote its TopK set as \( \mathcal{Z}_{\text{high}}^t \).

As shown in the lower part of Figure~\ref{fig:steer_experiment_main}, during the experiment, the \( \mathcal{Z}_{\text{high}} \) in the original activation \( \mathbf{H}_n \) for all tokens \( N \) are zeroed out. The steering latents \( \mathcal{Z}_{\text{high}}^t \) from \( d^t \) are then injected. Intuitively, if these steering latents carry meaningful information, the model's output may shift toward answering \( a^t \) for question \( q^t \), even though the input is question \( q \). We repeat the same experiment for BottomK.

\paragraph{Implementation Details.}
We primarily conduct experiments using the SAE from the Gemma Scope series~\cite{lieberum2024gemma}, trained on the 9th layer of the Gemma 2 9B Instruct model. For results related to the impact of layer choice, please refer to Appendix~\ref{appendix:different_layer}. Additionally, in Appendix~\ref{appendix:different_sae}, we extend our experiments to an SAE trained on the LLaMA 3 model to evaluate the generalizability of our method. Full implementation details are provided in Appendix~\ref{appendix:implementation_detail}.

\subsection{Perturbation Experiment Analysis}
Table~\ref{tab:first_experiment} upper section presents the Exact Match (EM) and F1 scores, comparing the performance of masking TopK and BottomK latent sets under both the Baseline and our proposed GradSAE method. The results are reported across values of \( K \in \{1, 10, 20, 30, 50\%\} \), as defined earlier. As shown in the ``w/o Task'' column, the initial LLM performance without any perturbation is 100\%.

Across all \( K \) settings, masking the TopK latents identified by both Baseline and GradSAE leads to a drop in EM and F1 scores, while masking BottomK latents maintains near-perfect performance. For example, both methods yield F1 scores around 99\% across \( K = 10, 20, 30, 50\% \) when masking BottomK latents. This confirms our \textbf{RQ1} hypothesis that not all activated latents equally contribute to output construction, with TopK latents having far greater influence.

Comparing GradSAE and the Baseline, masking GradSAE's TopK latents results in a more substantial performance drop. For instance, with \( K=1 \), GradSAE yields 80.45\% EM and 82.67\% F1, decreasing to 30.58\% EM and 43.33\% F1 at \( K = 50\% \). This consistent degradation highlights that GradSAE more precisely identifies latents critical to output, while the Baseline shows only moderate degradation. These results support \textbf{RQ2}, demonstrating GradSAE's superior effectiveness in identifying influential latents.

\subsection{Local Steering Experiment Analysis}

As shown in Table~\ref{tab:first_experiment} lower section ``w/o Task'' column, the initial LLM performance without local steering is 0\%. After applying local steering, both \textbf{Baseline} and \textbf{GradSAE} exhibit steering effects when the original TopK latents are replaced with those from a different question sharing the same context. GradSAE consistently outperforms the Baseline, especially at \( K=10 \), achieving a steering F1 score of 10.21\%. This supports \textbf{RQ3}, demonstrating that GradSAE-identified latents can steer the model toward answering a different question even when the input remains unchanged. However, as \( K \) increases, the steering effect diminishes. This is because replacing more TopK latents degrades output coherence as SAE reconstruction becomes less accurate. This aligns with findings from the first experiment, confirming that TopK latents are crucial for output construction, and excessive modification leads to output collapse.

Masking and replacing BottomK latents results in negligible steering across all \( K \) values (near 0\% EM and F1). Deeper analysis shows that local steering BottomK latents leaves the model's output unchanged. This is consistent with earlier results showing that BottomK latents have minimal influence. These findings confirm that modifying non-influential latents does not meaningfully steer or disrupt the output. For detailed statistical analysis, see Appendix~\ref{appendix:statistic_analysis}.

\section{Related Work}
SAEs have shown great promise in interpreting LLMs \cite{cunningham2023sparse, bricken2023monosemanticity, gao2025scaling}. However, most existing work focuses solely on input activations and assumes that these activations have a causal influence on the model's output \cite{templeton2024scaling, o2024steering}. In contrast, our work challenges this assumption. A detailed discussion on related work is provided in Appendix~\ref{appendix:related_work}.

\section{Conclusions}

In this work, we revisited the problem of identifying and local steering activated latents in SAEs for LLMs. We proposed two key hypotheses: (1) not all activated latents equally affect output, and (2) only highly influential latents are effective for steering. Through a series of experiments, we demonstrated that GradSAE more accurately identifies influential latents and enables more reliable steering, with results generalizing across different SAEs. 



\section*{Limitations}
In this paper, we focus on SAEs trained on instruction-tuned LLMs. While GradSAE is theoretically applicable to any SAEs, LLMs, and tasks, this study focuses on one of the most fundamental capabilities of modern LLMs with GradSAE, i.e., instruction following for question answering. 
Extending GradSAE to a broader range of scenarios, including but not limited to pre-trained LLMs, is an exciting direction for future work.

\section*{Acknowledgments}
Ninghao Liu is supported by the National Science Foundation (NSF) Grant \#2223768 and \#2507128.
Mengnan Du is supported by National Science Foundation (NSF) Grant \#2310261. The
views and conclusions in this paper are those of the authors and should not be interpreted as representing any funding agencies.


\bibliography{custom}

\begin{thebibliography}{31}
\providecommand{\natexlab}[1]{#1}

\bibitem[{Arora et~al.(2018)Arora, Li, Liang, Ma, and Risteski}]{arora2018linear}
Sanjeev Arora, Yuanzhi Li, Yingyu Liang, Tengyu Ma, and Andrej Risteski. 2018.
\newblock Linear algebraic structure of word senses, with applications to polysemy.
\newblock \emph{Transactions of the Association for Computational Linguistics}, 6:483--495.

\bibitem[{Ayonrinde(2024)}]{ayonrinde2024adaptive}
Kola Ayonrinde. 2024.
\newblock Adaptive sparse allocation with mutual choice \& feature choice sparse autoencoders.
\newblock \emph{arXiv preprint arXiv:2411.02124}.

\bibitem[{Bloom et~al.(2024)Bloom, Tigges, Duong, and Chanin}]{bloom2024saetrainingcodebase}
Joseph Bloom, Curt Tigges, Anthony Duong, and David Chanin. 2024.
\newblock Saelens.
\newblock \url{https://github.com/jbloomAus/SAELens}.

\bibitem[{Braun et~al.(2025)Braun, Taylor, Goldowsky-Dill, and Sharkey}]{braun2025identifying}
Dan Braun, Jordan Taylor, Nicholas Goldowsky-Dill, and Lee Sharkey. 2025.
\newblock Identifying functionally important features with end-to-end sparse dictionary learning.
\newblock \emph{Advances in Neural Information Processing Systems}, 37:107286--107325.

\bibitem[{Bricken et~al.(2023)Bricken, Templeton, Batson, Chen, Jermyn, Conerly, Turner, Anil, Denison, Askell, Lasenby, Wu, Kravec, Schiefer, Maxwell, Joseph, Hatfield-Dodds, Tamkin, Nguyen, McLean, Burke, Hume, Carter, Henighan, and Olah}]{bricken2023monosemanticity}
Trenton Bricken, Adly Templeton, Joshua Batson, Brian Chen, Adam Jermyn, Tom Conerly, Nick Turner, Cem Anil, Carson Denison, Amanda Askell, Robert Lasenby, Yifan Wu, Shauna Kravec, Nicholas Schiefer, Tim Maxwell, Nicholas Joseph, Zac Hatfield-Dodds, Alex Tamkin, Karina Nguyen, and 6 others. 2023.
\newblock Towards monosemanticity: Decomposing language models with dictionary learning.
\newblock \emph{Transformer Circuits Thread}.
\newblock Https://transformer-circuits.pub/2023/monosemantic-features/index.html.

\bibitem[{Bussmann et~al.(2024)Bussmann, Leask, and Nanda}]{bussmann2024batchtopk}
Bart Bussmann, Patrick Leask, and Neel Nanda. 2024.
\newblock Batchtopk sparse autoencoders.
\newblock \emph{arXiv preprint arXiv:2412.06410}.

\bibitem[{Chalnev et~al.(2024)Chalnev, Siu, and Conmy}]{chalnev2024improving}
Sviatoslav Chalnev, Matthew Siu, and Arthur Conmy. 2024.
\newblock Improving steering vectors by targeting sparse autoencoder features.
\newblock \emph{arXiv preprint arXiv:2411.02193}.

\bibitem[{Cunningham et~al.(2023)Cunningham, Ewart, Riggs, Huben, and Sharkey}]{cunningham2023sparse}
Hoagy Cunningham, Aidan Ewart, Logan Riggs, Robert Huben, and Lee Sharkey. 2023.
\newblock Sparse autoencoders find highly interpretable features in language models.
\newblock \emph{arXiv preprint arXiv:2309.08600}.

\bibitem[{Durmus et~al.(2024)Durmus, Tamkin, Clark, Wei, Marcus, Batson, Handa, Lovitt, Tong, McCain, Rausch, Huang, Bowman, Ritchie, Henighan, and Ganguli}]{durmus2024steering}
Esin Durmus, Alex Tamkin, Jack Clark, Jerry Wei, Jonathan Marcus, Joshua Batson, Kunal Handa, Liane Lovitt, Meg Tong, Miles McCain, Oliver Rausch, Saffron Huang, Sam Bowman, Stuart Ritchie, Tom Henighan, and Deep Ganguli. 2024.
\newblock \href {https://anthropic.com/research/evaluating-feature-steering} {Evaluating feature steering: A case study in mitigating social biases}.

\bibitem[{Elhage et~al.(2022)Elhage, Hume, Olsson, Schiefer, Henighan, Kravec, Hatfield-Dodds, Lasenby, Drain, Chen, Grosse, McCandlish, Kaplan, Amodei, Wattenberg, and Olah}]{elhage2022superposition}
Nelson Elhage, Tristan Hume, Catherine Olsson, Nicholas Schiefer, Tom Henighan, Shauna Kravec, Zac Hatfield-Dodds, Robert Lasenby, Dawn Drain, Carol Chen, Roger Grosse, Sam McCandlish, Jared Kaplan, Dario Amodei, Martin Wattenberg, and Christopher Olah. 2022.
\newblock \href {https://transformer-circuits.pub/2022/toy_model/index.html} {Toy models of superposition}.
\newblock \emph{Transformer Circuits Thread}.

\bibitem[{Feng et~al.(2018)Feng, Wallace, Grissom~II, Iyyer, Rodriguez, and Boyd-Graber}]{feng2018pathologies}
Shi Feng, Eric Wallace, Alvin Grissom~II, Mohit Iyyer, Pedro Rodriguez, and Jordan Boyd-Graber. 2018.
\newblock Pathologies of neural models make interpretations difficult.
\newblock In \emph{Proceedings of the 2018 Conference on Empirical Methods in Natural Language Processing}, pages 3719--3728.

\bibitem[{Galichin et~al.(2025)Galichin, Dontsov, Druzhinina, Razzhigaev, Rogov, Tutubalina, and Oseledets}]{galichin2025have}
Andrey Galichin, Alexey Dontsov, Polina Druzhinina, Anton Razzhigaev, Oleg~Y Rogov, Elena Tutubalina, and Ivan Oseledets. 2025.
\newblock I have covered all the bases here: Interpreting reasoning features in large language models via sparse autoencoders.
\newblock \emph{arXiv preprint arXiv:2503.18878}.

\bibitem[{Gao et~al.(2025)Gao, la~Tour, Tillman, Goh, Troll, Radford, Sutskever, Leike, and Wu}]{gao2025scaling}
Leo Gao, Tom~Dupre la~Tour, Henk Tillman, Gabriel Goh, Rajan Troll, Alec Radford, Ilya Sutskever, Jan Leike, and Jeffrey Wu. 2025.
\newblock \href {https://openreview.net/forum?id=tcsZt9ZNKD} {Scaling and evaluating sparse autoencoders}.
\newblock In \emph{The Thirteenth International Conference on Learning Representations}.

\bibitem[{Ghilardi et~al.(2024)Ghilardi, Belotti, and Molinari}]{ghilardi2024efficient}
Davide Ghilardi, Federico Belotti, and Marco Molinari. 2024.
\newblock Efficient training of sparse autoencoders for large language models via layer groups.
\newblock \emph{arXiv preprint arXiv:2410.21508}.

\bibitem[{He et~al.(2025)He, Zhao, Qiao, Yang, Payani, Ma, and Du}]{he2025saif}
Zirui He, Haiyan Zhao, Yiran Qiao, Fan Yang, Ali Payani, Jing Ma, and Mengnan Du. 2025.
\newblock Saif: A sparse autoencoder framework for interpreting and steering instruction following of language models.
\newblock \emph{arXiv preprint arXiv:2502.11356}.

\bibitem[{Joshi et~al.(2025)Joshi, Dittadi, Lachapelle, and Sridhar}]{joshi2025identifiable}
Shruti Joshi, Andrea Dittadi, S{\'e}bastien Lachapelle, and Dhanya Sridhar. 2025.
\newblock Identifiable steering via sparse autoencoding of multi-concept shifts.
\newblock \emph{arXiv preprint arXiv:2502.12179}.

\bibitem[{Lieberum et~al.(2024)Lieberum, Rajamanoharan, Conmy, Smith, Sonnerat, Varma, Kram{\'a}r, Dragan, Shah, and Nanda}]{lieberum2024gemma}
Tom Lieberum, Senthooran Rajamanoharan, Arthur Conmy, Lewis Smith, Nicolas Sonnerat, Vikrant Varma, J{\'a}nos Kram{\'a}r, Anca Dragan, Rohin Shah, and Neel Nanda. 2024.
\newblock Gemma scope: Open sparse autoencoders everywhere all at once on gemma 2.
\newblock \emph{arXiv preprint arXiv:2408.05147}.

\bibitem[{Marks et~al.(2024)Marks, Paren, Krueger, and Barez}]{marks2024enhancing}
Luke Marks, Alasdair Paren, David Krueger, and Fazl Barez. 2024.
\newblock Enhancing neural network interpretability with feature-aligned sparse autoencoders.
\newblock \emph{arXiv preprint arXiv:2411.01220}.

\bibitem[{Mudide et~al.(2024)Mudide, Engels, Michaud, Tegmark, and de~Witt}]{mudide2024efficient}
Anish Mudide, Joshua Engels, Eric~J Michaud, Max Tegmark, and Christian~Schroeder de~Witt. 2024.
\newblock Efficient dictionary learning with switch sparse autoencoders.
\newblock \emph{arXiv preprint arXiv:2410.08201}.

\bibitem[{O'Brien et~al.(2024)O'Brien, Majercak, Fernandes, Edgar, Chen, Nori, Carignan, Horvitz, and Poursabzi-Sangde}]{o2024steering}
Kyle O'Brien, David Majercak, Xavier Fernandes, Richard Edgar, Jingya Chen, Harsha Nori, Dean Carignan, Eric Horvitz, and Forough Poursabzi-Sangde. 2024.
\newblock Steering language model refusal with sparse autoencoders.
\newblock \emph{arXiv preprint arXiv:2411.11296}.

\bibitem[{Rajamanoharan et~al.(2024{\natexlab{a}})Rajamanoharan, Conmy, Smith, Lieberum, Varma, Kram{\'a}r, Shah, and Nanda}]{rajamanoharan2024improving}
Senthooran Rajamanoharan, Arthur Conmy, Lewis Smith, Tom Lieberum, Vikrant Varma, J{\'a}nos Kram{\'a}r, Rohin Shah, and Neel Nanda. 2024{\natexlab{a}}.
\newblock Improving dictionary learning with gated sparse autoencoders.
\newblock \emph{arXiv preprint arXiv:2404.16014}.

\bibitem[{Rajamanoharan et~al.(2024{\natexlab{b}})Rajamanoharan, Lieberum, Sonnerat, Conmy, Varma, Kram{\'a}r, and Nanda}]{rajamanoharan2024jumping}
Senthooran Rajamanoharan, Tom Lieberum, Nicolas Sonnerat, Arthur Conmy, Vikrant Varma, J{\'a}nos Kram{\'a}r, and Neel Nanda. 2024{\natexlab{b}}.
\newblock Jumping ahead: Improving reconstruction fidelity with jumprelu sparse autoencoders.
\newblock \emph{arXiv preprint arXiv:2407.14435}.

\bibitem[{Rajpurkar et~al.(2016)Rajpurkar, Zhang, Lopyrev, and Liang}]{rajpurkar2016squad}
Pranav Rajpurkar, Jian Zhang, Konstantin Lopyrev, and Percy Liang. 2016.
\newblock Squad: 100,000+ questions for machine comprehension of text.
\newblock In \emph{Proceedings of the 2016 Conference on Empirical Methods in Natural Language Processing}, pages 2383--2392.

\bibitem[{Scherlis et~al.(2022)Scherlis, Sachan, Jermyn, Benton, and Shlegeris}]{scherlis2022polysemanticity}
Adam Scherlis, Kshitij Sachan, Adam~S Jermyn, Joe Benton, and Buck Shlegeris. 2022.
\newblock Polysemanticity and capacity in neural networks.
\newblock \emph{arXiv preprint arXiv:2210.01892}.

\bibitem[{Shu et~al.(2025)Shu, Wu, Zhao, Rai, Yao, Liu, and Du}]{shu2025survey}
Dong Shu, Xuansheng Wu, Haiyan Zhao, Daking Rai, Ziyu Yao, Ninghao Liu, and Mengnan Du. 2025.
\newblock A survey on sparse autoencoders: Interpreting the internal mechanisms of large language models.
\newblock \emph{Findings of EMNLP 2025}.

\bibitem[{Soo et~al.(2025)Soo, Teng, Balaganesh, Guoxian, and YAN}]{soo2025interpretable}
Samuel Soo, Wesley Teng, Chandrasekaran Balaganesh, Tan Guoxian, and Ming YAN. 2025.
\newblock Interpretable steering of large language models with feature guided activation additions.
\newblock In \emph{ICLR 2025 Workshop on Building Trust in Language Models and Applications}.

\bibitem[{Taggart(2024)}]{Taggart}
Glen Taggart. 2024.
\newblock \href {https://www.alignmentforum.org/posts/HEpufTdakGTTKgoYF/prolu-a-nonlinearity-for-sparse-autoencoders#Learnable_parameters_of_a_sparse_autoencoder_} {Prolu: A nonlinearity for sparse autoencoders - ai alignment forum}.

\bibitem[{Templeton et~al.(2024)Templeton, Conerly, Marcus, Lindsey, Bricken, Chen, Pearce, Citro, Ameisen, Jones, Cunningham, Turner, McDougall, MacDiarmid, Freeman, Sumers, Rees, Batson, Jermyn, Carter, Olah, and Henighan}]{templeton2024scaling}
Adly Templeton, Tom Conerly, Jonathan Marcus, Jack Lindsey, Trenton Bricken, Brian Chen, Adam Pearce, Craig Citro, Emmanuel Ameisen, Andy Jones, Hoagy Cunningham, Nicholas~L Turner, Callum McDougall, Monte MacDiarmid, C.~Daniel Freeman, Theodore~R. Sumers, Edward Rees, Joshua Batson, Adam Jermyn, and 3 others. 2024.
\newblock \href {https://transformer-circuits.pub/2024/scaling-monosemanticity/index.html} {Scaling monosemanticity: Extracting interpretable features from claude 3 sonnet}.
\newblock \emph{Transformer Circuits Thread}.

\bibitem[{Wu et~al.(2024)Wu, Yao, Chen, Pan, Wang, Liu, and Yu}]{wu2024language}
Xuansheng Wu, Wenlin Yao, Jianshu Chen, Xiaoman Pan, Xiaoyang Wang, Ninghao Liu, and Dong Yu. 2024.
\newblock From language modeling to instruction following: Understanding the behavior shift in llms after instruction tuning.
\newblock In \emph{Proceedings of the 2024 Conference of the North American Chapter of the Association for Computational Linguistics: Human Language Technologies (Volume 1: Long Papers)}, pages 2341--2369.

\bibitem[{Wu et~al.(2025)Wu, Yuan, Yao, Zhai, and Liu}]{wu2025interpreting}
Xuansheng Wu, Jiayi Yuan, Wenlin Yao, Xiaoming Zhai, and Ninghao Liu. 2025.
\newblock Interpreting and steering llms with mutual information-based explanations on sparse autoencoders.
\newblock \emph{arXiv preprint arXiv:2502.15576}.

\bibitem[{Zhao et~al.(2024)Zhao, Devoto, Hong, Du, Gema, Wang, He, Wong, and Minervini}]{zhao2024steering}
Yu~Zhao, Alessio Devoto, Giwon Hong, Xiaotang Du, Aryo~Pradipta Gema, Hongru Wang, Xuanli He, Kam-Fai Wong, and Pasquale Minervini. 2024.
\newblock Steering knowledge selection behaviours in llms via sae-based representation engineering.
\newblock \emph{arXiv preprint arXiv:2410.15999}.

\end{thebibliography}

\newpage
\newpage

\appendix

\section{Proof of Gradiant Approximation}
\label{appendix:proof}
We aim to prove that the influence of latent activation $\mathbf{g}_{n,c}=p(\mathcal{Y}|\mathbf{H})-p(\mathcal{Y}|\mathbf{H}_{n,/c})$ defined in Equation~\eqref{eq:influence} can be approximated by the gradient-based approach described in Equation~\eqref{eq:impirical}. 
To start with, we consider that $p(\cdot|\cdot)$ implemented by an LLM is continued over its input domain~\cite{wu2024language}. 
Thus, we could extend $p(\mathcal{Y}|\mathbf{H})$ around the $\mathbf{H}_{n,/c}$ with the First-order Taylor expansion:
\begin{equation}
   \begin{aligned}
    p(\mathcal{Y}|\mathbf{H})\approx &\,\, p(\mathcal{Y}|\mathbf{H}_{n,/c})\,\, + \\ &\frac{\partial p(\mathcal{Y}|\mathbf{H})}{\partial \mathbf{H}}\Bigg|_{\mathbf{H}_{n,/c}}(\mathbf{H}-\mathbf{H}_{n,/c}).
\end{aligned} 
\label{eq:taylor}
\end{equation} 
Note that, since the only difference between $\mathbf{H}$ and $\mathbf{H}_{n,/c}$ is the latent activation at the $c$-th latent on the $n$-th word, i.e., $\mathbf{H}_{n,c}$, Equation~\eqref{eq:taylor} can be simplified as:
\begin{equation}
    p(\mathcal{Y}|\mathbf{H})\approx p(\mathcal{Y}|\mathbf{H}_{n,/c})  + \frac{\partial p(\mathcal{Y}|\mathbf{H})}{\partial \mathbf{H}_{n,c}} \mathbf{H}_{n,c}.
\end{equation}
Bringing this simplified form to the definition of influence $\mathbf{g}_{n,c}$ in Equation~\eqref{eq:influence}, we have $\mathbf{g}_{n,c}\approx \frac{\partial p(\mathcal{Y}|\mathbf{H})}{\partial \mathbf{H}_{n,c}} \mathbf{H}_{n,c}$. 
To this end, the influence of latent activations can be approximated with this gradient-based approach.

\begin{table}[t]
\centering
\caption{Datasets Statistics (Avg. = Average, \#Ex. = Number of Examples)}
\label{tab:dataset}
\begin{tabular}{lcc}
\toprule
DATASET & \multicolumn{2}{c}{SQuAD} \\ \hline
\multicolumn{1}{l}{}             & Train       & Valid       \\ \hline
\multicolumn{1}{l}{Context Avg. Length}  & 119.76     & 123.95      \\ \hline
\multicolumn{1}{l}{Question Avg. Length} & 10.06       & 10.22       \\ \hline
\multicolumn{1}{l}{Answer Avg. Length}   & 3.16        & 3.02        \\ \hline
\multicolumn{1}{l}{Avg. Questions / Context}  & 4.64     & 5.11      \\ \hline
\multicolumn{1}{l}{\#Ex.}        & 87.6k       & 10.6k  \\  
\bottomrule
\end{tabular}
\end{table}

\section{General Settings}
\subsection{Dataset}
\label{appendix:dataset}
\begin{figure}[t]
    \centering
    \includegraphics[width=0.90\linewidth]{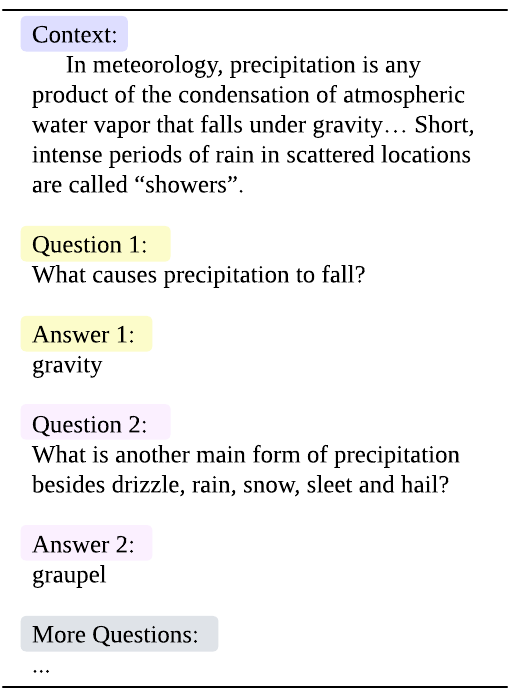}
    \caption{SQuAD dataset example.}
    \label{fig:data_example}
\end{figure}

In this paper, we use the SQuAD dataset~\cite{rajpurkar2016squad}, where each example consists of a context passage, a question, and an answer. Detailed dataset statistics are provided in Table~\ref{tab:dataset}. Each context is associated with multiple questions, on average around five per context. The answer to each question is a span extracted directly from the corresponding context. We choose this dataset for several reasons. First, the answers are short, averaging around three words, which makes evaluation more straightforward. Second, the context passages are sufficiently long, averaging about 120 words, which helps activate a diverse set of latent variables in the SAEs. This allows our GradSAE method to better identify the most critical latents from among multiple activated ones. Since our GradSAE is a training-free approach, we use only the validation set to evaluate its performance. We have shown an example of the data in Figure \ref{fig:data_example}.

\subsection{Metrics}
\label{appendix:metrics}
We adopt the standard SQuAD evaluation metrics: Exact Match (EM) and token-level F1. EM measures the percentage of predictions that exactly match a ground-truth answer, while F1 captures the overlap at the token level between predictions and references. To ensure fair and consistent evaluation, we normalize text by lowercasing, removing punctuation, trimming extra whitespace, and excluding uninformative words such as ``a'', ``an'', and ``the''. These two metrics together offer a balanced and comprehensive assessment of span-level prediction quality.

\section{Implementation Details}
\label{appendix:implementation_detail}
Since our experimental design involves instruction-based question answering, in theory, our GradSAE method can be applied to any SAE model supported by the SAE-Lens \cite{bloom2024saetrainingcodebase}. However, in practice, we find that only instruction-tuned LLMs are capable of successfully answering questions during our experiments. Consequently, in the main paper, we primarily conduct experiments using the SAE from the Gemma Scope series~\cite{lieberum2024gemma}. Specifically, we use the gemma-scope-9b-it-res-canonical SAE, which is trained on activations from the 9th layer of the Gemma 2 9B Instruct model. This SAE features an overcomplete latent space with 131,000 dimensions. Additionally, in Appendix~\ref{appendix:different_sae}, we extend our experiments to an SAE trained on the LLaMA 3 8B Instruct model to evaluate the generalizability of our method across different instruction-tuned LLMs. All experiments are performed on an A100 SXM4 GPU with 80 GB of memory. For experiments involving randomness, we fix the random seed at 42 to ensure reproducibility. Additionally, we evaluate our approach on the same SAEs trained on activations from the 20th and 31st layers of the Gemma 2 9B Instruct model.

\begin{table}[ht]
\centering
\caption{Empirical analysis of SAE activations. ``Avg.'' denotes the average number of non-zero latent activations per input.}
\label{tab:empricial_study}
\scalebox{0.95}{
\begin{tabular}{lcc}
\toprule
 & Baseline & GradSAE \\ \toprule
Activation Avg. & \multicolumn{2}{c}{103.41} \\ \hline
50\% Avg. & \multicolumn{2}{c}{51.46} \\ \hline
Cross TopK Overlap & \multicolumn{2}{c}{18.06\%} \\ \hline
Cross BottomK Overlap & \multicolumn{2}{c}{0.33\%} \\ \hline
Inner TopK Overlap & 89.91\% & 50.46\% \\ \hline
Inner BottomK Overlap & 92.17\% & 0.15\% \\ \hline
\bottomrule
\end{tabular}}
\end{table}

\section{More Experiment Results}
\label{appendix:more_results}

\subsection{Statistic Analysis}
\label{appendix:statistic_analysis}

In this section, we empirically analyze and compare the activated latents identified by the \textbf{Baseline} and \textbf{GradSAE}. As shown in Table~\ref{tab:empricial_study}, both approaches yield, on average, approximately 103 non-zero latent activations per example in the SAE activation. The average value of 50\% (defined in Section \ref{sec:first_experiment_design}) is approximately 51. To measure the overlap between the two methods, we compute the cross overlap between the TopK (K = 50\%) sets selected by Baseline and GradSAE. We observe that, on average, only 18.06\% of the latents in the Baseline's TopK set are also present in GradSAE's TopK set. This suggests that many latents with high input-side activation may not correspond to high output-side gradients. This highlights input activation alone does not reliably indicate causal influence on the output.

Interestingly, the cross overlap between the BottomK sets of Baseline and GradSAE is even lower, around 0.33\%. This suggests that only a few low-activation latents in the Baseline may still carry some gradient signal, whereas other majority latents might have zero gradient. Despite this minimal overlap, Table~\ref{tab:first_experiment} upper section shows that masking either method's BottomK latents has almost no negative impact on performance. This implies that the vast majority of latents are non-influential to the model's output, and the number of truly uninfluential latents is significantly greater than the value of ``50\%''. 

Other than the cross overlap, we also measure the inner TopK and BottomK latent overlap (with K = 50\%) across different prompts sharing the same context. Specifically, since each context in our dataset corresponds to multiple questions, we calculate the overlap of activated latents across these different questions within the same context. As shown in Table \ref{tab:empricial_study}, the baseline approach exhibits a high degree of overlap: 89.91\% for TopK latents and 92.17\% for BottomK latents. In contrast, GradSAE shows much lower overlaps: 50.46\% for TopK and only 0.15\% for BottomK. This discrepancy is expected, as the baseline activations are purely input-driven and reflect prompt-level similarity. Since prompts with the same context are textually similar except for minor changes in the question, their activations tend to be similar as well. However, GradSAE's activations are guided by both the input and the output gradient. This gradient signal selectively filters out latents that do not influence the final prediction, leading to more diverse and dynamic activation patterns. These findings further reinforce our earlier results, demonstrating that GradSAE is more effective in identifying truly influential latents that contribute to model output.

\begin{table}[t]
\centering
\caption{Results of the perturbation experiment using SAEs trained on different layers of the LLM. This table evaluates how the choice of layer affects the effectiveness of GradSAE and Baseline.}
\label{tab:layer_experiment}
\scalebox{0.85}{
\begin{tabular}{cccccc}
\toprule
 & & \multicolumn{2}{c}{Layer 20} & \multicolumn{2}{c}{Layer 31} \\ \hline
 &  & EM & F1 & EM & F1 \\ \hline
\multirow{2}{*}{Baseline} & TopK & 73.83 & 82.53 & 69.52 & 77.69 \\
 & BottomK & 96.14 & 99.39 & 96.73 & 99.68  \\ \hline
\multirow{2}{*}{GradSAE} & TopK & 9.06 & 38.35 & 6.53 & 17.38 \\
 & BottomK & 96.81 & 99.74 & 96.73 & 99.80 \\
\bottomrule
\end{tabular}}
\end{table}

\begin{table*}[t]
\centering
\caption{Results from the perturbation experiment using SAE trained on the LLaMA 3 8B Instruct model. This table evaluates the generalization ability of GradSAE.}
\label{tab:different_sae_experiment}
\begin{tabular}{cccccccc}
\toprule
 & & \multicolumn{2}{c}{K = 1} & \multicolumn{2}{c}{K = 10} & \multicolumn{2}{c}{K = 50\%}\\ \hline
 &  & EM & F1 & EM & F1 & EM & F1 \\ \hline
\multirow{2}{*}{Baseline} & TopK & 88.57 & 95.97 & 80.02 & 88.73 & 65.71 & 78.94 \\
 & BottomK & 100.0 & 100.0 & 97.14 & 99.78 & 97.14 & 99.78 \\ \hline
\multirow{2}{*}{GradSAE} & TopK & 65.71 & 79.24 & 58.14 & 72.98 & 48.57 & 60.99 \\
 & BottomK & 100.0 & 100.0 & 97.14 & 99.78 & 97.14 & 99.78 \\
\bottomrule
\end{tabular}
\end{table*}

\subsection{Different Layer}
\label{appendix:different_layer}

While the main experiment focused on an SAE trained on layer 9 of Gemma 2 9B Instruct model, \citet{lieberum2024gemma} also developed SAEs for layers 20 and 31. To investigate the impact of the choice of layer on GradSAE, we repeated our first experiment using SAEs from these different layers. Specifically, we compared the effects of masking the TopK versus BottomK latents for both Baseline and GradSAE, using K = 50\%. The results, presented in Table \ref{tab:layer_experiment}, reveal two key findings. First, consistent with our previous results (Table \ref{tab:first_experiment} upper section), masking the TopK latents harms model performance more than masking BottomK latents for both methods across all layers (9, 20, 31). This reinforces our previous idea that not all activated latents have the same influence on the model output.

Second, we observe differing trends between the methods across layers. For the Baseline method, the performance gap between TopK and BottomK masking decreasing at deeper layers (comparing layer 9 vs. 20 vs. 31). This suggests that latents activated solely on input becomes less effective at influencing outputs in later layers. However, for GradSAE, the significant performance gap between TopK and BottomK masking is maintained across all layers. This highlights GradSAE's robustness to layer choice, likely because it selects latents based on both input activation and output gradients. These results further validate GradSAE's ability to identify latents truly influential to the model's output.

\subsection{Different SAE}
\label{appendix:different_sae}

In addition to experiments on the Gemma 2 9B Instruct model, we also evaluated our first experiment setup using an SAE trained on the LLaMA 3 8B Instruct model. Specifically, we use the llama-3-8b-it-res-jh SAE, which was trained on activations from the 25th layer and features a latent space with 65,536 dimensions. In this setting, we set \( K \in \{1, 10, 50\%\} \), where \( K=50\% \) corresponds to approximately 26 latents, given that the llama-3-8b-it-res-jh SAE activates around 52 latents per token on average. As shown in Table~\ref{tab:different_sae_experiment}, the performance trends are consistent with our main experiment that masking the TopK latents identified by GradSAE results in a substantial drop in model performance, whereas masking the BottomK latents has negligible impact. While the Baseline method also exhibits this general trend, the performance gap between TopK and BottomK masking is notably smaller compared to GradSAE. These results further support \textbf{RQ2}, demonstrating that GradSAE can effectively identify the most influential latents even when applied to different SAEs trained on different instruction-tuned LLMs.

\section{Related Works}
\label{appendix:related_work}

\subsection{Sparse AutoEncoder}
SAEs have emerged as a widely used and highly promising tool for interpreting the internal mechanisms of LLMs \cite{cunningham2023sparse, bricken2023monosemanticity, gao2025scaling, rajamanoharan2024jumping}. An SAE is a neural network framework designed to learn an overcomplete and sparse representation of model activations, which helps disentangle the superimposed features within LLMs \cite{elhage2022superposition}. This directly addresses the polysemanticity problem, where a single neuron responds to multiple unrelated concepts. Traditionally, training a SAE involves balancing reconstruction fidelity with strong sparsity constraints, ensuring that only a small subset of latents activate for any given input. As a result, SAEs can extract more interpretable, monosemantic features, offering a clearer and more human-understandable view of LLM internal behaviors.

Beyond the traditional SAE, a variety of SAE variants have been proposed to further enhance interpretability, improve reconstruction quality, or optimize training efficiency. These advancements can be broadly categorized into two areas: architectural improvements and training strategy improvements \cite{shu2025survey}. In terms of architectural improvements, models such as Gated SAE \cite{rajamanoharan2024improving}, TopK SAE \cite{gao2025scaling}, Batch TopK SAE \cite{bussmann2024batchtopk}, ProLU SAE \cite{Taggart}, JumpReLU SAE \cite{rajamanoharan2024jumping}, and Switch SAE \cite{mudide2024efficient} introduce modifications to the activation mechanisms (e.g., TopK selection or gated activations) to better enforce sparsity and refine feature selection. On the other hand, improvements in training strategies include approaches like Layer Group SAE \cite{ghilardi2024efficient}, Feature Choice SAE \cite{ayonrinde2024adaptive}, Mutual Choice SAE \cite{ayonrinde2024adaptive}, Feature Aligned SAE \cite{marks2024enhancing}, and End-to-end SAE \cite{braun2025identifying}. These methods enhance feature selection, alignment, and training efficiency while preserving the core architecture of traditional SAEs. In this paper, GradSAE introduces a training-free approach that leverages output gradients falling under architectural improvements.

\subsection{Model Steering}
SAEs have emerged as a powerful tool not only for interpreting the internal mechanisms of LLMs but also for steering their behavior, since SAEs can identify distinct, human-interpretable features within LLMs. Once these features are identified, interventions can be performed by activating or suppressing the corresponding latents during inference. Several recent studies have explored and enhanced the use of SAEs for steering LLMs. SAIF \cite{he2025saif} proposes a framework for interpreting instruction-following capabilities in LLMs by identifying instruction-relevant SAE features and demonstrates how manipulating these features can effectively steer instruction-following behavior. SAE-TS \cite{chalnev2024improving} introduces a method to construct steering vectors that precisely target specific SAE features while minimizing unintended side effects, leading to more controlled and coherent model outputs. SpARE \cite{zhao2024steering} leverages SAE representations to detect and resolve context-memory knowledge conflicts at inference time, enabling LLMs to selectively use contextual or parametric knowledge. Mutual Information-Based Explanations (MIE) \cite{wu2025interpreting} addresses frequency bias in SAE feature interpretations and proposes runtime steering strategies that adjust feature activations based on more meaningful, discourse-level explanations. FGAA \cite{soo2025interpretable} further refines activation steering by optimizing over SAE latents, creating highly targeted and interpretable steering vectors that improve steering effectiveness while maintaining output quality. SSAEs \cite{joshi2025identifiable} propose a new unsupervised method that learns sparse, identifiable latent representations of multi-concept shifts, enabling accurate concept-level steering without requiring curated supervision. However, these approaches implicitly assume a direct correspondence between latents activated solely by the input and the aspects of the output they aim to steer. In contrast, GradSAE challenges this assumption by arguing that the model's output must also be considered when identifying influential latents for steering. By incorporating output-side gradient information, GradSAE provides a more accurate attribution of which latents truly drive model outputs, leading to more effective and reliable steering interventions.

\end{document}